\documentclass{article}

\usepackage{arxiv}

\usepackage[utf8]{inputenc} 
\usepackage[T1]{fontenc}    
\usepackage{hyperref}       
\usepackage{url}            
\usepackage{booktabs}       
\usepackage{amsfonts}       
\usepackage{nicefrac}       
\usepackage{microtype}      
\usepackage{lipsum}
\usepackage{graphicx}
\graphicspath{ {./images/} }

\title{A `Sourceful' Twist: Emoji Prediction Based on Sentiment, Hashtags and Application Source}

\author{ Pranav Venkit\textsuperscript{1}\thanks{These authors contributed equally.}
, Zeba Karishma\textsuperscript{2}\footnotemark[1]
, Chi-Yang Hsu\textsuperscript{1}\footnotemark[1]
, Rahul Katiki\textsuperscript{1}\footnotemark[1],
\\ \textbf{Kenneth Huang\textsuperscript{1}, 
        Shomir Wilson\textsuperscript{1}, Patrick Dudas\textsuperscript{1}} \\
\\
\textsuperscript{1}College of IST, \textsuperscript{2}Department Of CSE \\
Pennsylvania State University \\
\texttt{{pranav.venkit,zbk5052,cxh5437,rbk5315,txh710,shomir,pmd19}@psu.edu} \\}

\begin{document}
\maketitle
\begin{abstract}
We widely use emojis in social networking to heighten, mitigate or negate the sentiment of the text. Emoji suggestions already exist in many cross-platform applications but an emoji is predicted solely based a few prominent words instead of understanding the subject and substance of the text. Through this paper, we showcase the importance of using Twitter features to help the model understand the sentiment involved and hence to predict the most suitable emoji for the text. Hashtags and Application Sources like Android, etc. are two features which we found to be important yet underused in emoji prediction and Twitter sentiment analysis on the whole. To approach this shortcoming and to further understand emoji behavioral patterns, we propose a more balanced dataset by crawling additional Twitter data, including timestamp, hashtags, and application source acting as additional attributes to the tweet. Our data analysis and neural network model performance evaluations depict that using hashtags and application sources as features allows to encode different information and is effective in emoji prediction.

\end{abstract}

\section{Introduction}

With the increase in usage of Emojis in social media, it has not only changed the way we communicate with each other, but has also provided us with sentiment rich data in large scale to perform research.
These pictorial characters allow a user to describe objects, situations, and reactions.
The meaning of various emojis used have been investigated through social media platforms such as Twitter, with the aid of just the text present in the tweet \cite{shiha2017effects,eisner2016emoji2vec}.
Despite the emojis pervasive use, only a few studies have explored non-textual features associated with a tweet such as hashtags or the device used \cite{Felbo_2017,davidov2010enhanced}. However, these play an important role in both influencing the way a person tweets, and the sentiment conveyed by the tweet. The primary motivation of our research, hence, is to develop an Artificial Intelligence system that is able to predict emojis by contributing to better natural language understanding through the aid of not just text but also other tweet features. 



The goal of our task stems from the SemEval 2018 task 2, Multilingual Emoji Prediction~\cite{barbieri2018semeval}, which focuses on predicting the emoji for two different languages, {\em i.e.}, English and Spanish.
Given the top 20 emojis as class labels, the task is to predict most likely emoji associated with a given text-only Twitter message.
In this work, we perform an in-depth analysis of this problem. Our analysis pinpoints two major shortcomings of the existing methods and the dataset. First, the Semeval emoji dataset has a significant unequal class distribution. While some classes occupied the majority of the data, most classes only accounted for less than 3\% of the data. Second, given the original dataset only provides textual data, we believe the textual information alone cannot provide sufficient information for prediction. To address the listed shortcomings, this paper proposes two following solutions: (1) To further understand emoji behavioral patterns, we crawl additional Twitter data to balance the class distribution as well as to include timestamp, hashtags, and application source as additional attributes to the tweet. (2) To extend the prediction of the emojis in tweets by additionally analysing hashtags and application sources. Our work focuses on just the English dataset.
We will release the tweet IDs of the complete dataset and open source the code used for crawling, creating a normalized Twitter dataset and feature extraction for future exploration of the emoji prediction task.

\section{Related Work}

The study on the usage and semantics of emoji has led to many variations of the novel emoji prediction task. Due to the numerous emojis currently present for our usage, their cross-device implementations and the increase in the level of ambiguity created by cultural differences, it has become a problem in sentiment analysis \cite{hallsmar2016multi}. An evaluation performed by Miller \cite{miller2016blissfully} portrays a notable difference in a user's perception of the actual meaning of an emoji within and between various platforms. Chen et al \cite{DBLP:journals/corr/abs-1807-07961} embedded each emoji into two distinct vectors to get positive and negative-sense embeddings for each emoji to deal with semantic ambiguity. Kimura and Katsurai \cite{10.1145/3110025.3110139} extracted sentiment words from WordNet-Affect and calculated the co-occurrence frequency between the sentiment words and each emoji, based on those ratios and sentiment words from WordNet, assigned each emoji with a multi-dimensional vector whose elements indicate the strength of the corresponding sentiment.
Guibon et al. \cite{guibon:hal-01529708} used SentiStrength\footnote{http://sentistrength.wlv.ac.uk/} to calculate the sentiment polarities and ESR \cite{10.1371/journal.pone.0144296} to compute the emoji polarities and compared them to find out the contextually right emoji label.

The contribution closest to ours used hashtags. Davidov \cite{davidov2010enhanced} utilized hashtags to create training data, but they limited their experiments to sentiment and non-sentiment classification rather than performing the multilabel emotion classification presented in Wolny \cite{Wolny2016EmotionAO}, where hashtags were used as labels alongside selected emojis.
Felbo et al. \cite{Felbo_2017}, employed emojis as sentence representation to get better understanding on how emojis can not only be used as effective labels but also to read the emotion, sentiment, and sarcasm present in a sentence.
The novel feature for our work will be to assess how attributes associated with a tweet such as application source and hashtags impact the usage of emojis, their meanings. The greater goal will be to see if these attributes can be used as features to perform an environmentally aware emoji prediction. 
For this purpose, we have utilized the 20 most frequent emoji list provided by Barbieri \cite{semeval2018task2} as our class labels.


\section{Preprocessing and Data Analysis}

\subsection{Data}

\begin{figure}
    \includegraphics[width=9cm, height=4cm]{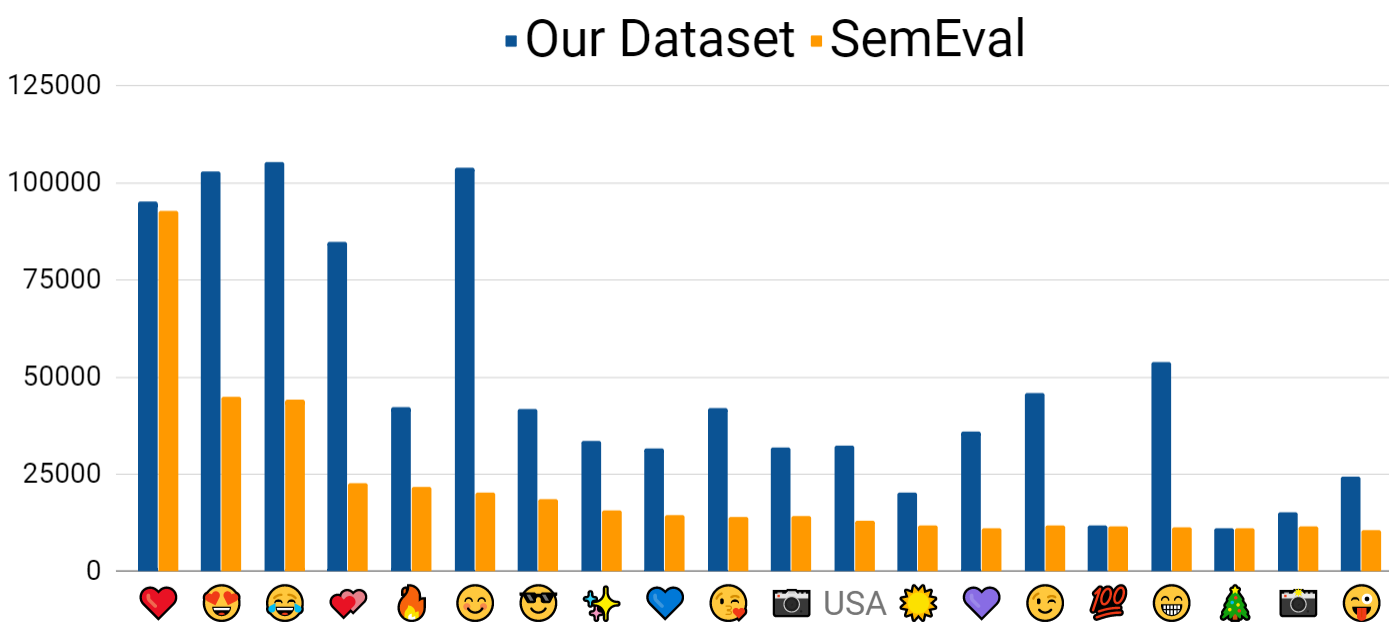}
    \centering
    \caption{Comparative Analysis of Label Distribution for SemEval and Our Preprocessed Dataset}
    \label{fig:data2}
\end{figure}

The data, provided by the SemEval 2018 Task 2 \cite{semeval2018task2}, consists of 500k tweets in the English language. These were geolocalized to the United States between October 2015 to February 2017. \cite{semeval2018task2}.
However, our analysis on the dataset showed an unequal class distribution among top 20 most used emojis. While some classes occupied the majority of the data, most classes only accounted for less than 3\% of the data.
We therefore crawled for more English tweets from December 2019 to May 2020 using Tweepy\footnote{https://github.com/tweepy/tweepy}.
The crawled data provides us with a larger dataset with tweets from a broader period of time. We crawled a total of 1.98 million tweets. However, after cleaning and removal of retweets, the dataset was down-sampled to 1.1 million tweets. A comparative analysis of label distribution of SemEval and our dataset is shown in Figure \ref{fig:data2}. We plan on releasing this new dataset publicly as a modified corpus for future investigation of the emoji prediction task  and to act as a platform for future NLP research.

\subsection{Preprocessing}

The first step of preprocessing was to remove emojis, remove links and anonymize user mentions using emoji\footnote{https://github.com/carpedm20/emoji} library which was followed by the removal of retweets and null strings from the tweet dataset. We use  Ekphrasis\footnote{https://github.com/cbaziotis/ekphrasis} \cite{baziotis2017datastories} for preprocessing, which is designed to clean social media-oriented data. Through this, we performed tokenization, spell correction and word segmentation (for splitting hashtags). We were also able to identify and replace special strings such as emoticons, dates, acronyms and censored data. To normalize the tweets, we converted the text to lowercase and removed URLs, user handles and the special characters. This aided in refining the vocabulary size without losing essential sentiment or linguistic information. Finally, we extracted the hashtags as a separate feature, by performing a similar flow of preprocessing. 

\subsection{Analysis}
\begin{figure}
    \includegraphics[width=7.5cm, height=3.5cm]{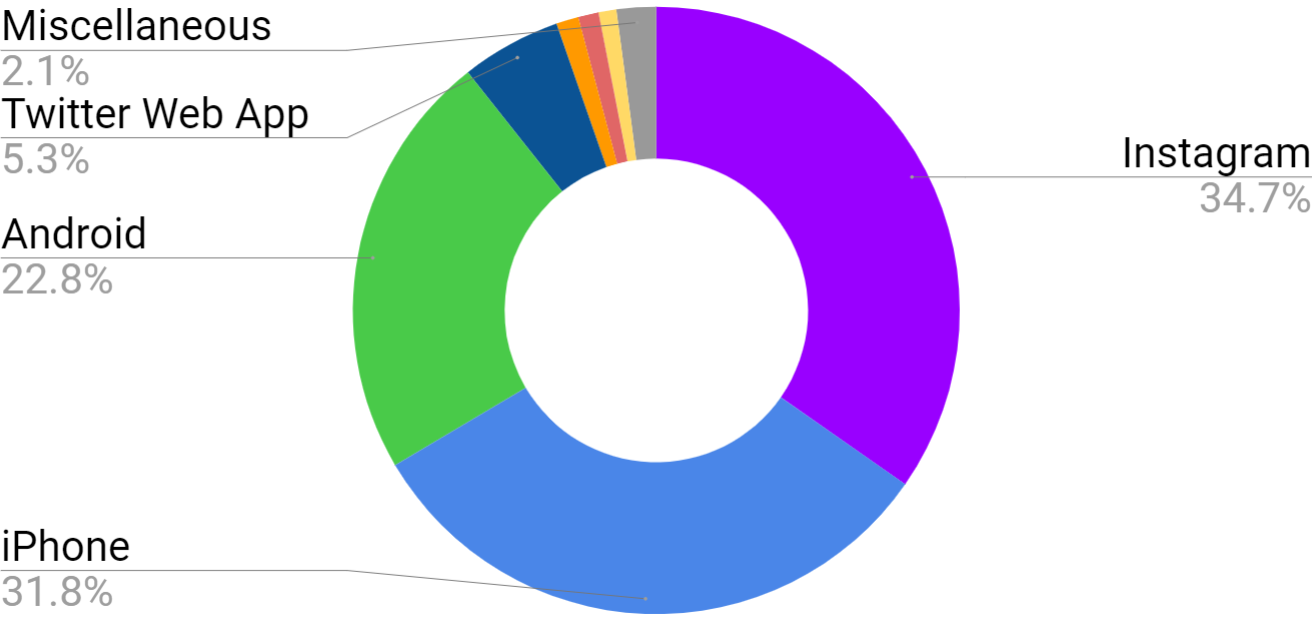}
    \hspace{\fill}
    \includegraphics[width=8cm, height=4cm]{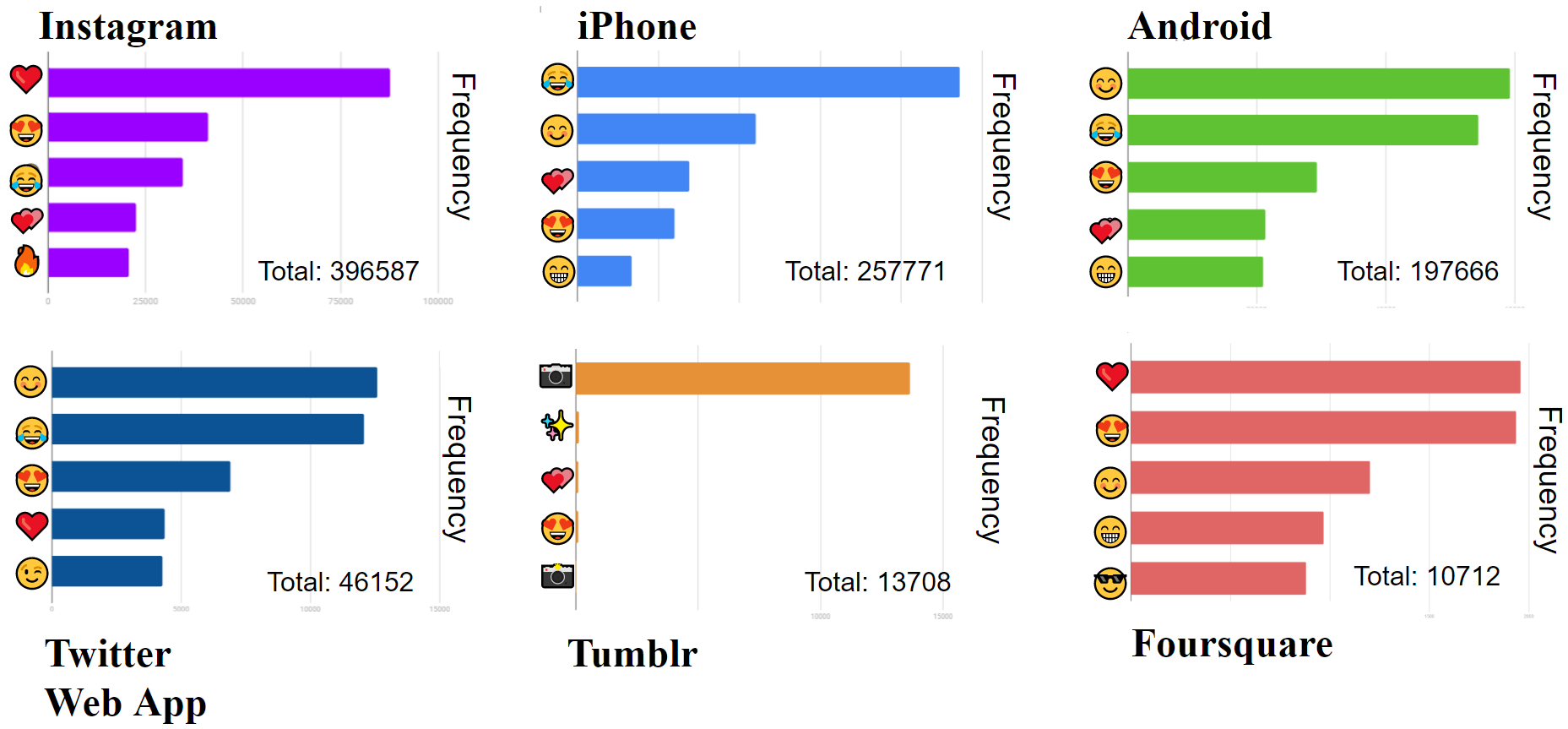}
    \centering
    \caption{Source Distribution of Tweets \& Top 5 Emoji Analysis of the Top 6 Application Source}
    \label{fig:source2}
\end{figure}

Our approach aims to understand the effective change in performance by considering the tweet attributes as an important and separate feature during classification.
First, we obtain the top features from both the tweets and the extracted hashtags of the complete dataset. The intention of doing this was to understand whether texts or hashtags can provide a good distinction to the semantic meaning of an emoji.
To obtain word-level text analysis, we implement conventional Random Forest classifier
and used the scikit-learn package to procure top word features for each label/emoji, as shown in Table \ref{table:1}. 
From this table it can be comprehended as to how texts and hashtags, individually and combined, provide a linguistic understanding of each emoji without pre-trained knowledge.
\begin{table}[h!]
\centering
\begin{tabular}{|c|c|c|} 
 \hline
 Label & Text & Hashtags \\ [0.5ex] 
 \hline
 \includegraphics[scale=.4]{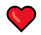} & love, family, happy, day, park  & love, family, day, life, memorial \\ 
 \hline
\includegraphics[scale=.4]{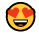} & beautiful, love, cute, gorgeous, product & month, love, life, insecure, hbo \\
 \hline
 \includegraphics[scale=.4]{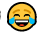} & lol, lmao, thank, love, shit & uyajola, anonymous, funny, sundays, lol,  \\
 \hline
 \includegraphics[scale=.4]{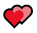} & love, thank, safe, stay, happy, & fettaho, nur, love, engineer, lives  \\
 \hline
 \includegraphics[scale=.4]{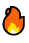} & lit, hot, new, album, heat & lit, turn, deals, nbs, music \\ 
\hline
\end{tabular}
\caption{Top 5 Feature Extracted for Text and Hashtag Analysis of the Top 5 Emojis}
\label{table:1}
\end{table}
Next, we extract the application source variable of the complete dataset. 
We have depicted the various source of tweets in Figure \ref{fig:source2}. We identify the topmost source to be Instagram, followed by Twitter from iPhone and Android devices respectively.
From segregating and analyzing each emoji with the source, we identify how different sources have different preferences and usage of emojis. This has been well explained by Tauch et al. \cite{tauch2016roles} where they demonstrate how the most used emoji in one platform in not the same in another. It is interesting to also note how platforms such as Tumblr have the lower occurring emoji, i.e "\includegraphics[scale=.4]{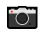}", to be most used, due to its possible artistic inclination of use or the format of which tweets are posted from this platform. The other more possible reason for the change in usage of emojis can be explained by the appearance of emojis in each platform \cite{hallsmar2016multi} and the heuristic placement of the emoji for different device and application screen.
Hence, due to the unique behavior of the emojis with sources and hashtag, we conclude on using these Twitter features as a potential addition in the prediction of the emoji.

\section{Experimentation and Results}

\begin{table}[h!]
\centering
\begin{tabular}{||c|c|c|c|c|c|c||}
 \hline\hline
 Model & 
 Setting & Acc & Prec & Rec & F1 
 \\ [0.5ex] 
 \hline \hline
 RandomForest 
 & Semeval & 0.2836 & 0.2320 & 0.1902 & 0.1929\\
 LinearSVC 
 & Semeval &  0.3268 & \textbf{0.3309} & 0.2099 & 0.2014\\
 
 BiLSTM & 
 Semeval  &0.3138 &  0.1466 & 0.1820 &0.1466 
 \\
 BiLSTM& 
 Complete Dataset  & 0.3220 & 0.2655 & 0.2223 & 0.2042
 \\
 \hline\hline
& 
 Text + Hashtags  & 0.3065 & 0.2560 & 0.2124 & 0.2033 
 \\
 BiLSTM & 
 Text + Source & \textbf{0.3372} & 0.2814 & \textbf{0.2301} & 0.2110 
 \\
& 
Text + Hashtags + Source & 0.2966 & 0.2545 & 0.2225 & \textbf{0.2139} 
\\

 \hline \hline
\end{tabular}
\caption{Accuracy, macro precision, macro recall, and macro F1 various feature/dataset combinations.
}
\label{table:2}
\end{table}
We use six variations of the dataset, consisting of different combinations of feature embedding. This is shown in Table \ref{table:2}. The upper table depicts the effect of using different datasets, and the lower table shows the effect of using different feature combinations. With this comparative examination, we demonstrate how these selected Twitter features help change the prediction at hand. We also show the behavioural insights of an emoji in a tweet by perceiving the change in the pattern of auto-metrics. For the classifier, in order to evaluate the effectiveness of the additional features, we implemented a simple yet effective deep learning model, Bidirectional LSTM (BiLSTM) encoder-decoder model. In which, each additional feature is converted to an embedding and concatenated to the text embedding. Following the competition's main expectation: improve the performance of less frequent emojis, we're emphasizing on the macro precision, macro recall, and macro F1 as listed in Table ~\ref{table:2}. 



The first insight from the result, as shown in Table \ref{table:2}, is the effectiveness of the crawled dataset in accordance to the SemEval provided dataset. For both the model, we see that our crawled dataset results in greater score in prediction. This demonstrated the effect of increasing the dataset to a more normalized one. The second insight is the drastic increase in performance of the model while using the "application source" as a feature as compared to "hashtags". This is noticed in both the models. Therefore this result demonstrates the influence an application source can have in the usage of an emoji, irrespective of its absolute linguistic meaning. We also provide a better performing model by using both the hashtag and application source as additional features. The greater performance was demonstrated by the BiLSTM model that was developed to include the embedding layer of all these features.

\section{Conclusion}

In this paper, we present a multifaceted analysis on multiple features of a tweet and showcase their importance using machine learning and graphical representation. Our final result depicts how "application source" proves to be a more effective feature for predicting an emoji for a tweet. The best performing model was achieved through the addition of all the tweet feature, including hashtags. To complement and encourage further research on emoji analysis using the addition of such features, we have released two variations of our dataset consisting of a total of 1.98 million tweet IDs, and the code to crawl these tweets and their attributes \footnote{github.com/PranavNV/PretzelXD}. 
For the next stage, we intend to include an attention layer \cite{vaswani2017attention} to our model, to try and achieve a greater outlook on the important words and hashtags in a tweet.
We also propose to include both the polarity and the subjectivity values of the tweets, through TextBlob \cite{loria2014textblob} library, to understand and predict the effectiveness of such a feature in emoji prediction. As this project's direction is inclined on just English, we plan on performing the same direction of prediction for the Spanish dataset as well.

\bibliographystyle{unsrt} 
\bibliography{coling2020}



\end{document}